\documentclass[sigconf]{acmart}
\usepackage{makecell}
\usepackage{graphicx}
\usepackage{pdfpages}
\usepackage[utf8]{inputenc}
\usepackage{array}
\usepackage{wrapfig}
\usepackage{multirow}
\usepackage{tabularx}
\usepackage{algorithm}
\usepackage{algpseudocode}
\usepackage{amsmath}

\AtBeginDocument{
  \providecommand\BibTeX{{
    \normalfont B\kern-0.5em{\scshape i\kern-0.25em b}\kern-0.8em\TeX}}}

\setcopyright{acmcopyright}
\copyrightyear{2020}
\acmYear{2020}
\acmDOI{XXX.XXX/XXXXX.XXXXXX}

\acmConference[Conference 'XX]{Conference 'XX: ACM International Conference on XXXXX }{XXXX XX--XX , 2020}{XXXX, XXXX}
\acmBooktitle{Conference 'XX: ACM International Conference on XXXXX, October 19--23, 2020, XXXX, XXXXX}
\acmPrice{15.00}
\acmISBN{XXX-X-XX-XXXX-X/XX/XX}

\begin{document}

\title{Revisiting Regex Generation for Modeling Industrial Applications by Incorporating Byte Pair Encoder}

\author{Desheng Wang}
\authornote{Both authors contributed equally to this research.}
\authornote{Corresponding author.}
\email{desheng.wangds@antfin.com}
\affiliation{
  \institution{Ant Financial Services Group}
  \city{Hangzhou}
  \country{China}
}

\author{Jiawei Liu}
\email{muyan.ljw@antfin.com}
\authornotemark[1]
\affiliation{
  \institution{Ant Financial Services Group}
  \city{Hangzhou}
  \country{China}
}

\author{Xiang Qi}
\email{ qixiang.qx@antfin.com}
\affiliation{
  \institution{Ant Financial Services Group}
  \city{Hangzhou}
  \country{China}
}

\author{Baolin Sun}
\email{ xuanfeng.sbl@antfin.com}
\affiliation{
  \institution{Ant Financial Services Group}
  \city{Hangzhou}
  \country{China}
}

\author{Peng Zhang}
\email{ minghua.zp@antfin.com}
\affiliation{
  \institution{Ant Financial Services Group}
  \city{Hangzhou}
  \country{China}
}

\begin{abstract}
Regular expression is important for many natural language processing tasks especially when used to deal with unstructured and semi-structured data. 
This work focuses on automatically generating regular expressions and proposes a  novel genetic algorithm to deal with this problem.
Different from the methods which generate regular expressions from character level, we first utilize byte pair encoder (BPE) to extract some frequent items, which are then used to construct regular expressions. 
The fitness function of our genetic algorithm contains multi objectives and is solved based on evolutionary procedure including crossover and mutation operation. 
In the fitness function, we take the length of generated regular expression,  the maximum matching characters and samples for positive training samples, and the minimum matching characters and samples for negative training samples into consideration.
In addition, to accelerate the training process, we do exponential decay on the population size of the genetic algorithm. 
Our method together with a strong baseline is tested on 13 kinds of challenging datasets. The results demonstrate the effectiveness of our method, which outperforms the baseline on 10 kinds of data and achieves nearly \textbf{9.7} percent  improvement on F1 score on average.
By doing exponential decay, the training speed is approximately \textbf{100} times faster than the methods without using exponential decay. In summary, our method possesses both effectiveness and efficiency, and can be implemented for the industry application.
\end{abstract}



\maketitle

\section{Introduction}
Regular expression, which can also be abbreviated as regex, utilizes a sequence of characters to define a search pattern and has been investigated for a long time.
Based on \emph{Nondeterministic Finite Automaton} (NFA) and \emph{Deterministic Finite Automaton} (DFA) \cite{rabin1959finite}, it can efficiently extract some important information such as bank card numbers and Chinese certificate numbers from unstructured and semi-structured data such as raw text.
Due to the effectiveness and flexibility, regular expression has been widely used in the areas of Natural Language Processing (NLP) \cite{manning1999foundations}, Data Mining (DM) \cite{han2011data}, Information Retrieval (IR) \cite{manning2008introduction}, etc.
To assure the quality of the regular expressions constructed for some specific tasks,  however, expertise with respect to regular expressions is necessary, which  means  the construction of regular expressions is of great difficulty.

\begin{figure}
    \centering
    \includegraphics[width=0.75\linewidth]{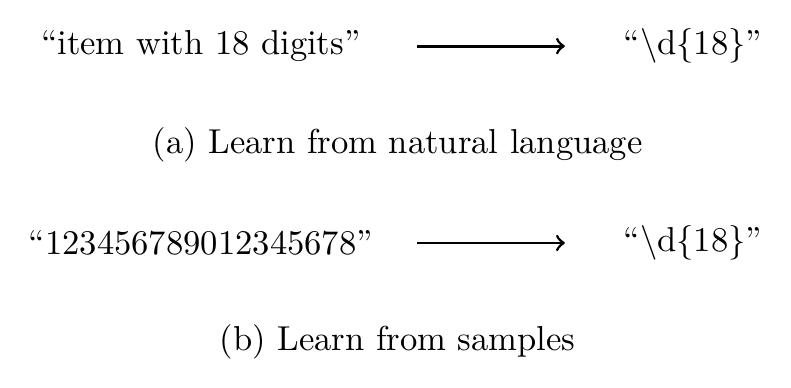}
    \caption{Auto regex generation tasks. (a) stands for  generating regex from natural language, and (b) represents learning regex from samples. }
    \label{fig:task_category}
\end{figure}

To overcome the drawbacks of generating regular expressions, some regular expression generation methods are proposed. Generally, these methods can be classified into two categories.
As Figure \ref{fig:task_category} shows, the picture (a) stands for the method that regular expressions are generated from natural language.
To address the problem, \cite{LocascioNeural} constructed a end-to-end model \cite{sutskever2014sequence} based on the Long Short-Term Memory Neural Network \cite{hochreiter1997long} and trained the model by utilizing a synthetic parallel corpus of natural language descriptions and regular expressions.
Although the method in \cite{LocascioNeural} achieved the state-of-the-art performance, the authors in \cite{zhong2018semregex} suggested that the distinct characteristics between the synthetic datasets and the real-world datasets might affect the ability of the end-to-end model which was demonstrated to achieve extremely low effectiveness on real-world data.
Hence, a large amount of real-world parallel data is necessary to guarantee the effectiveness of the end-to-end model.
In addition, deep models will consume a huge amount of computation resources and can hardly be applied in some low-resource environments.

The other kind of methods construct regular expressions from samples of the desired behavior as the picture (b) shows.
\cite{Bartoli2016Inference} treated the problem as a program synthesis task and dealt with it by implementing an evolutionary procedure.
Based on the structure of trees, they first generated some candidate regular expressions by utilizing templates.
Next, a genetic algorithm containing crossover and mutation operations was used to find the result with the best fitness.
Although this method was proved to be effective in \cite{Bartoli2016Inference}, our experimental results show that their algorithm is time-consuming and can hardly deal with more than 5000 training samples in a tolerable time.  Due to the computational complexity, this method is impossible to be applied to model some industrial applications.
In general, these methods mentioned above generate regular expressions from character-level, which ignore the correlations between characters in the training sample.

In this work, we revisit the task of constructing regular expression from samples for modeling industrial applications.
In order to improve both effectiveness and efficiency, we modify the genetic algorithm mentioned in \cite{Bartoli2016Inference} and propose three improvements.
First of all, we modify the original fitness function and remove the limitation of forcing the genetic algorithm to choose the shortest regular expression.
We find that the limitation will make regular expressions too generalized (e.g. the regex ``.*" ) to distinguish the positive samples from those adversarial samples.
In this work, the regular expression whose length is similar to the positive training samples' length is recognized to be the best.
Secondly, a compressed algorithm named byte pair encoder (BPE) \cite{sennrich2015neural}\cite{gage1994new} was introduced to extract some frequent items from training corpus so that we can make use of these frequent items to construct more specific regular expressions.
Furthermore, with the help of BPE, the searching space is shrunk and the genetic algorithm can be more efficient than before.
Last but not the least, motivated by the simulated annealing algorithm \cite{van1987simulated}, we do exponential decay on the population size of the genetic algorithm  for accelerating the training process.
In reality, positive samples may contain some noise. In this work, according to the analysis of the datasets, we assume the percentage of noise will not surpass 5\%.
In order to improve the robustness of our method, we do divide-and-conquer and make our method focus on the incorrectly matched samples. When the percentage of the remained training samples is lower than 5\%, we will stop the training process.

Our method and a strong baseline in \cite{Bartoli2016Inference} are trained on both positive samples and negative samples. The learned regular expressions are expected to match the positive samples and mismatch the negative samples.
In experiment, our method and the baseline are evaluated on 13 kinds of challenging datasets including Chinese certificate number, mobile phone number, email, etc. The results indicate the effectiveness of our method, which outperforms the baseline by approximately 50 percent on average. Especially, our method even achieves nearly 30 percent on the international mobile equipment identity (IMEI), car engine number and Chinese certificate number.
By doing exponential decay, the training speed is nearly 100 times faster than the method without using exponential decay.
Our contributions are summarized as follows.
\begin{itemize}
    \item To the best of our knowledge, we are the first to introduce the BPE algorithm into regular expression generation. By utilizing frequent items, the generated regular expressions are much more specific. Hence, when giving some challenging negative samples, the results show much better performance. The experimental results indicate that our method approximately achieves 9.7 percent improvement on F1 score on average.
    \item We modify the fitness function and remove the limitation of forcing the genetic algorithm to choose the shortest regular expression. We argue that the regular expression with the maximum matching characters for positive training samples and the minimum matching character for negative training samples and having similar length to the positive training samples is considered to be the best.
    \item By doing exponential decay on population, the training speed is nearly 100 times faster than before, which makes the regex generation available for modeling industrial applications.
\end{itemize}
The rest of the thesis is organized as follows.
We will introduce some related work about this thesis in Section 2.
Section 3 lists the basic ideas of our method.
We describe the experiments in Section 4.
In this section, we introduce the experiment settings, result analysis, hyperparameter analysis and procedure to conduct these experiments.
We will give the conclusion of our work and how these can be implemented into future work in the last section.

\section{Related Work}
\textbf{Automatical Regex Generation (ARG)}
With the surges of data volume, more and more semi-structured and unstructured data are generated.
Regular Expression as a long-researched topic can efficiently extract some useful information from these data.
Due to the effectiveness, regular expressions have been widely used in many natural language tasks, e.g. name entity extraction (NER) \cite{chiu2016named} ,  information retrieval \cite{manning2008introduction, li2008regular, bartoli2017active}.
However, the construction of  regular expressions is hard, tedious and  expertise-demanded. To address this problem, some automatic regex generation methods are proposed.
In general, regular expression generation can be divided into two categories including regular expressions from natural language and regular expressions from samples. In the first category, regular expressions are learned from some descriptions written in  natural language.  \cite{ranta1998multilingual} developed a rule-based system that defines a natural language interface to regular expression generation. Recently, \cite{LocascioNeural} treated this task as a machine translation task. Based on sequence to sequence framework, they made use of a long short term memory neural network to address the problem, in which the input of encoder was natural language and the output of decoder was regular expression.  However, this end-to-end model suffers from demanding a lot of parallel training data, which is resource-cost.

The second category directly learns regular expressions  from training samples which can be easily obtained. \cite{prasse2015learning} developed a method to generate the regular expression for recognizing email. However, their method can not be easily modified for other kinds of data. 
In \cite{Bartoli2016Inference, Bartoli2015Learning, bartoli2014playing}  treated the problem as a program synthesis task and proposed a genetic algorithm to deal with it.
Based on syntactic trees, they first generated some candidates by utilizing templates.  The leaf nodes denote the regular expression grammar and the non-leaf nodes are operators including concatenate operator, group operator, etc. The regular expression is generated by implementing a deep-first algorithm on the syntactic tree. 
In order to find optimal regular expression,  they defined a fitness function which was used to evaluate the quality of candidate regular expression. During the searching process, an evolutionary procedure was carried out including crossover and mutation operations.  Although this method is effective, it is time-consuming and costs lots of resources to find optimal solution.

\textbf{Byte Pair Encoder (BPE)}
These methods mentioned above are character-based, which means regular expressions are generated character by character. We argue that character-based methods can hardly capture some vital connections between characters.
For instance, in a Chinese certificate number whose length is 18,  the characters from 7 to 10 indicate the year of birth. Hence, characters from 7 to 8 can only be "19" or "20". However, character-based methods ignore this limitation and the generated regular expression will be ``\textbackslash d\textbackslash d\textbackslash d\textbackslash d\textbackslash d\textbackslash d\textbackslash d\textbackslash d\textbackslash d\textbackslash d\textbackslash d\textbackslash d\textbackslash d\textbackslash d\textbackslash d\textbackslash d\textbackslash d\textbackslash w", which can easily be cheated by some fake examples.
To construct more specific regular expressions, this work proposes a frequent-items-base method.
Our method first utilizes BPE \cite{sennrich2015neural} \cite{gage1994new} to extract some frequent items from the training examples by iteratively replacing the most frequent item with some unseen tokens.
Then, these frequent items are exploited to construct regular expressions.
Recently, BPE as a word segmentation technique has been widely used in many NLP tasks such as machine translation and text classification, and has achieved some significant improvements.
In this thesis, we modified the original BPE algorithm to control the granularity of the generated frequent items.

\section{Methods}
As we mentioned above, the construction of regular expressions needs expertise and is difficult.
To deal with this problem, we develop a novel Genetic Algorithm (GA) to automatically generate regular expressions in this section.
GA is an evolutionary algorithm and utilizes simulated actions including crossover and mutation to find the solution with optimal fitness from searching space.

\subsection{Task Statement}
As we mentioned above, our task in this thesis is to automatically generated regular expressions from training samples.
Given positive samples set $P = \{ s_p^1, s_p^2, \cdots, s_p^k\}$ and negative samples $N = \{s_n^1, s_n^2, \cdots, s_n^t\}$, the task needs to find a regular expression, which can perfectly distinguish the positive samples from the negative samples from searching space $\zeta$. The  formula is defined in Equation \ref{task statement}.
\begin{equation}
    \label{task statement}
    res = \arg\max_{r \in \zeta}\  fitness(r)
\end{equation}
where the fitness function is to evaluate the quality of the generated regular expressions and is defined in Equation \ref{fitness function}.

\subsection{Fitness Function}
In a standard genetic algorithm, it is vital to define a fitness function to evaluate the quality of candidates. We treat the definition of fitness function as a multi-objective problem.
Apparently, a satisfied regular expression should match more positive samples and less negative samples. In addition, from character perspective, the length of matched substring in positive samples is the longer the better while the fewer characters matched in negative samples is the better. Last but not the least, the length of generated regular expression is taken into consideration. In \cite{Bartoli2016Inference}, the regular expression with fewer characters are considered to be the better one. However, in experiments, we find that those regular expressions with the shortest length are too generalized to distinguish the positive sample from negative samples.
In this thesis, we argue that these regular expressions whose length is similar to the positive sample's length is the best.

For a regular expression $r$, given positive samples $P = \{ s_p^1, s_p^2, \\ \cdots, s_p^k\}$ and negative samples $N = \{s_n^1, s_n^2, \cdots, s_n^t\}$ , we first define an identifier function to denote if $r$ can totally match a sample $s \in P \cup N$.
\begin{equation}
\label{identifier}
match(r, s)=\left\{
\begin{aligned}
&1,\ when\ regex\ r \ totally\  matched \ s\\
&0,\ otherwise
\end{aligned}
\right.
\end{equation}
For regular expression $r$, the matched character number of the sample $s$ is defined as $count(r, s)$.

\begin{equation}
\label{fitness function}
     fitness(r) = P_s + P_c + L_{score}  \\
\end{equation}
where
\begin{equation}
\label{fitness function component}
    \begin{split}
        P_s =\  & \frac{\Sigma_{i \in P }\ match(r, i)}{\Sigma_{i \in P \cup N}\ match(r, i)} \\
        ~\\
            P_c =\  & \frac{\Sigma_{i \in P}\ match(r, i) * count(r, i)}{\Sigma_{i \in P \cup N}\ match(r, i) * count(r, i)} + \\
            & \frac{\Sigma_{i \in P } [1- match(r, i)] * count(r, i)}{\Sigma_{i \in P \cup N}\ [1- match(r, i)] * count(r, i)} \\
        ~\\
    L_{score} =\  & e^{-|\ len(r) - \frac{1}{k} \Sigma_{i \in P}\ len(i)\ |}
    \end{split}
\end{equation}
In Equation \ref{fitness function component}, function $len()$ denotes the length of string, $e$ is Euler's number.

\textbf{Tricks}
Obviously, calculation of the fitness function is time-consuming because of the large amount of regex matching operations. To accelerate the training process, we utilize two tricks to implement our algorithm.
\begin{itemize}
    \item When doing regex matching, we insert ``$\hat{\ }$" and ``\$" the at the head and tail of regular expression, respectively. This operation will greatly reduce the number of substrings to be matched.
    \item In each training epoch, we maintain a cache in our program to store the fitness score. For a regular expression, the fitness score will not be recalculated if we find the corresponding result in the cache.
\end{itemize}

\begin{figure}[!hbtp]
  	\begin{minipage}[t]{.5\textwidth}
  		\centering
      \includegraphics[width=.6\textwidth]{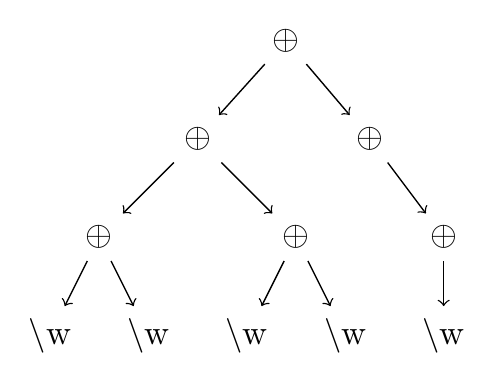}
  	\end{minipage}
  	\begin{minipage}[t]{.5\textwidth}
  	  \centering
       (a)
  	\end{minipage}
  	\\
  	\begin{minipage}[t]{.5\textwidth}
  		\centering
      \includegraphics[width=.4\textwidth]{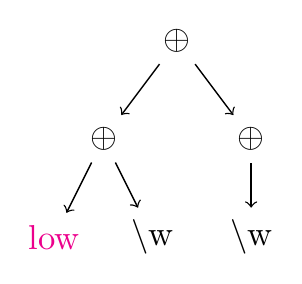}
  	\end{minipage}
  	\begin{minipage}[t]{.5\textwidth}
  	  \centering
       (b)
  	\end{minipage}
    \caption{Initialization of regex which is represented by as syntactic tree. Picture (a) is character-based strategy which is used in \cite{Bartoli2016Inference}. Our frequent item-based method is shown in picture (b).}
    \label{fig:initialization}
\end{figure}

\subsection{Initialization}
On the initialization stage of GA, some candidates will be randomly generated. Then, crossover and mutation will be conducted between these candidates to find the optimal result.
Motivated by \cite{Bartoli2016Inference}, in this thesis,  all regular expressions are constructed by using syntactic trees in which leave nodes are basic regular expression units chosen from the terminal set, and non-leaf nodes stand for operators including concatenation, matching one or more characters, etc.

The terminal sets are defined as follows.
  \begin{enumerate}
    \item Alphabet constants: ``a", ``b", . . . ,``y",``z”,``A",``B", . . . ,``Y" ,``Z";
    \item Digit constants:  ``0",``1", . . . ,``8",``9”;
    \item Symbols constants: ``.", ``:", ``,", ``;", ``\_", ``=", ``\textbackslash", ``'", ``\textbackslash\textbackslash", ``/", ``?", ``!", ``\}", ``\{", ``(", ``)", ``[", ``]", ``<", ``>", ``@", ``\#";
    \item Alphabet ranges and digit ranges: ``a-z", ``A-Z", ``0-9";
    \item Common character classes: ``\textbackslash w", ``\textbackslash d";
    \item Wildcard character: ``.";
  \end{enumerate}

The functional sets are defined as follows:
  \begin{enumerate}
    \item The concatenation operator ``$t_1t_2$". ``$\oplus$" denotes concatenation operator in Figure \ref{fig:initialization};
    \item The group operator ``($t_1$)";
    \item The list match operator``[t$_1$]" and the list not match operator ``[\^{}t$_1$]";
    \item The match one or more operator ``t$_1$++";
    \item The match zero or more operator ``t$_1$*+";
    \item The match zero or one operator ``t$_1$?+";
    \item The match min max operator ``t$_1$\{$n$,$m$\}+", $n$ is the minimum times, $m$ is the maximum times;
  \end{enumerate}

\begin{algorithm}[h]
   \caption{BPE: Byte Pair Encoding Algorithm with Dynamic Proportion Threshold Control}
   \label{alg::BPE}
   \begin{algorithmic}[1]
             \Require
               $p$: proportion threshold;\par
               $n$: the size of string samples in vocab
             \Ensure
               best: bpe tokens;
             \State import re
             \State \quad
             \State def pair\_freq\_stats(vocab):
             \State \quad pair2freq = collections.defaultdict(int)
             \State \quad for word, freq in vocab.items():
             \State \qquad              symbols = word.split()
              \State \qquad  \quad             for i in range(len(symbols) - 1):
              \State \qquad   \qquad                pair2freq[symbols[i], symbols[i + 1]] += freq
                \State  \quad    return pair2freq
              \State \quad
              \State  def merge(best, vocab\_in):
              \State \quad  vocab\_out = \{\quad\}
              \State \quad  bigram = re.escape(' '.join(best))
              \State \quad  patten = re.compile(r'(?<!\textbackslash S)' + bigram + r'(?!\textbackslash S)')
              \State \quad for word in vocab\_in:
              \State \qquad     word\_out = patten.sub(''.join(best), word)
              \State \qquad     vocab\_out[word\_out] = vocab\_in[word]
              \State \quad return vocab\_out
              \State \quad

        \State vocab = \{`l o w </w>': 5, `l o w e r </w>': 2, `n e w e s t </w>': \par
        \qquad 6, `w i d e s t </w>': 3\}
        \State percent = 1.0
        \State while percent $\ge$ p:
        \State \quad pair2freq = pair\_freq\_stats(vocab)
        \State \quad best = max(pair2freq, key=pair2freq.get)
        \State \quad percent = pair2freq.get(best)/$n$
        \State \quad if percent $\ge$ p:
        \State \quad \quad vocab = merge(best, vocab)
        \State \quad \quad print(best)
   \end{algorithmic}
\end{algorithm}

The final regular expression is generated by using the deep-first search algorithm on the corresponding syntactic tree.
As we mentioned above, however, traditional methods are character-based and suffer from low precision when training samples are insufficient. To solve this problem, in this thesis, we propose a frequent item-based initialization method. We first extract frequent items from training samples and then tokenize these samples by using BPE.
In Figure \ref{fig:initialization}, picture (a) denotes the initialization stage of traditional method and picture (b) illustrates ours.
The basic idea of BPE is to iteratively replace the most frequent item with some unseen tokens.  In this thesis, however, we find that these extremely  sophisticated frequent items will do harm to the generalization performance of the generated regular expressions. In the original BPE, the granularity of frequent items is controlled by the hyperparameter of training epochs. However, the experiments demonstrate that this strategy is unable to generate gratifying frequent items. Hence, we modify the original BPE and set a threshold to control the frequency of the most frequent items. The algorithm will be terminated once the frequency of corresponding frequent item is smaller than the threshold.
The pseudo code of the BPE algorithm is described in Algorithm \ref{alg::BPE}.


\begin{figure*}
    \centering
    \includegraphics[width=0.5\textwidth]{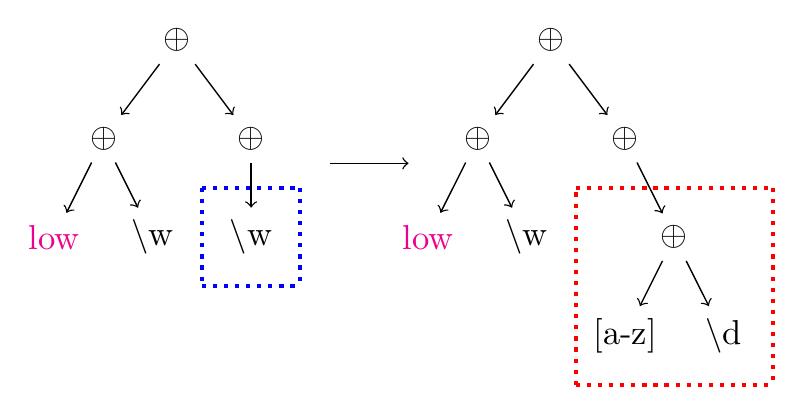}
    \caption{Mutation operation. This operation randomly replace a subtree with another tree. As shown in this figure, the original subtree in the blue dash box is replaced by the subtree in the red dash box.}
    \label{fig:mutation}
\end{figure*}

\subsection{Evolution}
During the initialization stage of the genetic algorithm, we randomly generate some  candidates, which are also named populations and evaluated based on the fitness function. Next, the operations of crossover and mutation are carried out on these populations to find the result with the best fitness in the searching space.

\textbf{Population Decay}
In \cite{Bartoli2016Inference}, the population size is invariant in each training epoch. However, the training speed is strongly connected with the population size. To improve the efficiency of our algorithm, we do exponential decay on population size. In each epoch, the population size $N_{pop}^i$ is defined as follows.
\begin{equation}
\label{identifier}
N_{pop}^i=\left\{
\begin{aligned}
&\max(N_{pop}^{i-1} * \lambda,\ N_{pop}^{min}),\  i \in [2, E]\\
&N_{pop},\ i=1
\end{aligned}
\right.
\end{equation}
where $\lambda, \lambda \in(0,1]$ is the decay parameter on population size, $E$ denotes epoch size, $N_{pop}$ is the initialized population size and $N_{pop}^{min}$ is the minimum number of population size during the training process and is used to prevent our algorithm from under-fitting.

\textbf{Mutation}
In mutation operation, for a regular expression, we randomly choose a subtree and replace it with another syntactic tree.
In the $i_{th}$ epoch, the mutation operation will be repeated for $0.1 * N_{pop}^i$ times to generate $0.1 * N_{pop}^i$ new populations. Details are shown in Figure \ref{fig:mutation}.

\begin{figure*}
    \centering
    \includegraphics[width=1.0\linewidth]{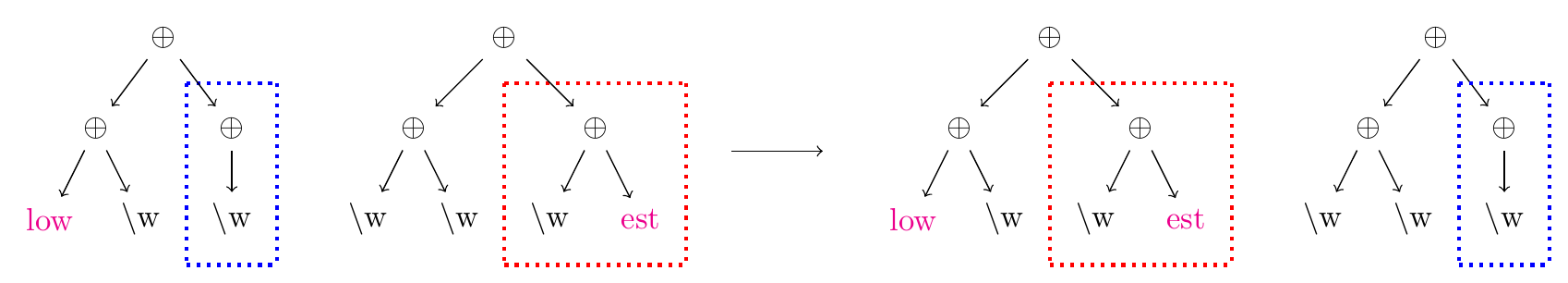}
    \caption{Crossover Operation. This Operation is used to exchange the subtrees of two different trees.}
    \label{fig:crossover}
\end{figure*}

\textbf{Crossover}
 In crossover operation, we randomly select two candidates from populations and switch their subtrees. In the $i_{th}$ epoch, the crossover operation will be repeated for $0.8 * N_{pop}^i$ times to generate $0.8 * N_{pop}^i$ new populations. The whole process of crossover is illustrated in Figure \ref{fig:crossover}.

To jump out of the local optimal result,  we randomly generate $0.1 * N_{pop}^i$ new populations for the  $i_{th}$ epoch. In the end, we choose the best $N_{pop}^i$ regular expressions as the output of the $i_{th}$ epoch according to the fitness function.

\begin{algorithm}[h]
  \caption{RGGB:Regex Generation based on GA and BPE}
  \label{alg::RGGB}
  \begin{algorithmic}[1]
    \Require
      $S=P \cup N$
      : $P$ is positive strings set , $N$ is negative strings set. \par
      $P = \{s_p^1,s_p^2,.....s_p^{k-1},s_p^k\}$, \par
      $N = \{s_n^1,s_n^2,.....s_n^{t-1},s_n^{t}\}$, \par
      $iter$: iteration for algorithm, \par
      $percent$: proportion threshold for bpe algorithm, \par
      $p_{div}$: precision threshold for divide and conquer, \par
      $it_{div}$: iteration threshold for divide and conquer. \par
    \Ensure
      $result$: regex which generated from $S$ can perfectly distinguish the positive samples from the negative samples.
    \State \quad
    \State initial $R=\varnothing$, iteration = 0, $iter_{best}$ = 0
    \State $bpe\_tokens$ = BPE($P$, $percent$)
    \State $R_{candidate}$ = init($P$, $bpe\_tokens$, $N_{pop}$)
      \Repeat
           \State $R_{c}$ =  crossover\_generate($R_{candidate}$, $N_{pop}*0.8$)
           \State $R_{m}$ = mutate\_generate($R_{candidate}$, $N_{pop}*0.1$)
           \State  $R_{r}$ = random\_generate($N_{pop}*0.1$)
           \State $R_{candidate}= R_{candidate} \cup R_{c}  \cup R_{m}  \cup R_{r} $
           \State $F= \varnothing$
           \For{each $r_i$ in $R_{candidate}$}
              \State $F = F \cup fitness(r_{i})$
           \EndFor
           \State $R_{candidate}$ = top\_K($R_{candidate}$, $N_{pop}$)
           \State iteration += 1
           \State best = get\_best($F$,$R_{candidate}$)
           \State $iter_{best}$ += 1
           \State $p_{best}$ = get\_precision($F$,best)
           \If { $p_{best}$  $\ge$  $p_{div}$ and $iter_{best}$   $\ge$  $it_{div}$}
             \State $R =  R \cup  best$
             \State $P$ = remove($P$, best)
             \State $R_{candidate}$ = init($P$, $bpe\_tokens$, $N$)
             \State $iter_{best}$ = 0
           \EndIf
       \Until iteration < $iter$
      \State $result$ = join(R)
  \end{algorithmic}
\end{algorithm}

\subsection{Divide and Conquer}
In reality, positive samples always contain some noise and may do harm to the performance of our algorithm.
In this work, according to the analysis of dataset, we assume that the percentage of noise will not surpass 5\%.
For robustness, after several training epochs, those regular expressions whose precision on training samples surpasses a predefined threshold will be selected into a candidate set and will not be utilized for training  anymore in the following training epochs.
Besides, positive samples which are correctly recognized by any regular expression in the candidate set will be removed in the next training epoch.
When the percentage of the remained training samples is lower than 5\%, we will stop the training process.
This strategy can make our algorithm focus on the incorrectly matched samples.
In the end, the result is reported by concatenating the regular expressions in the candidate set using symbol ``|" , which is shown in  Equation \ref{divide_and_conquer}.
\begin{equation}
\label{divide_and_conquer}
\begin{split}
Result = regex_1\ |\  regex_2\ |\  \cdots \ | \ regex_n
\end{split}
\end{equation}
where $n$ is the size of the candidate set. The core idea of our algorithm is summarized in Algorithm \ref{alg::RGGB}.

\subsection{Complexity Analysis}
For convenience, we reuse the notation defined in the previous subsections.
$N_{pop}$ denotes the size of population for syntax trees,
$k$ is the number of positive samples,
$t$ stands for the number of negative samples,
$\Delta$ denotes the complexity of a regex for matching a sample,
$\lambda$ refers to the exponential decay parameter on population size, and
$E$ is epoch size.

The cost of evolution for the $i_{th}$ epoch is $\Delta*(k+t)*N_{pop}*\lambda^{i}$, Hence the total complexity is defined in Equation \ref{complexity of our}
\begin{equation}
\label{complexity of our}
\begin{split}
C_{our} =\ & \Delta*(k+t)*N_{pop}*\lambda^{1} + \\ \ & \Delta*(k+t)*N_{pop}*\lambda^{2} + \cdots + \\ & \Delta*(k+t)*N_{pop}*\lambda^{E} \\
=\  & \Delta*(k+t)*N_{pop}*\frac{1-\lambda^E}{1-\lambda}
\end{split}
\end{equation}
Similarly, the computational complexity of the baseline is defined as follows.
\begin{equation}
\label{complexity of baseline}
\begin{split}
C_{baseline} = \Delta*(k+t)*N_{pop}*E
\end{split}
\end{equation}
Apparently, our method is much more efficient than the baseline.

\section{Experiments}
In this section, we first describe the construction of the dataset and give some experiment settings. To validate the effectiveness of our method, a strong baseline is introduced. Then, the results of experiment are discussed. At last, we analyze some vital hyperparameters, which may have a great effect on the performance of our method.

\begin{table}[h]
    \centering
    \renewcommand{\arraystretch}{1.5}
   	\begin{tabular}{p{0.1\columnwidth}p{0.45\columnwidth}p{0.25\columnwidth}}
         \hline
         \textbf{Order} & \textbf{Data Type}  & \textbf{Size} \\
         \hline
         1  & Mac Address   & 10000000  \\

         2  & IMEI   &  10000000  \\

        3  &  IP Address & 6499413 \\

        4  &  Invoice Code & 33606 \\ 

        5  &  Invoice Number & 10000000 \\ 

        6 &  Mobile Number & 10000000 \\

        7  &  House id & 1248507 \\

        8  &  Car Engine Number & 12599128 \\

        9  &  Company Unicode & 10000000 \\ 

        10  & Chinese Certificate Number & 10000000 \\

        11  &  Car License & 13739000 \\

        12  & Email & 10394457 \\

        13  &  Bankcard Number & 14071316 \\
        \hline
    \end{tabular}
    \caption{Details of Dataset.}
    \label{tab:dataset}
\end{table}

\subsection{Dataset and Experiment Settings}
To the best of our knowledge, there is no open-source dataset in the area of learning regular expressions for samples.
For comparison, we select 13 categories of challenging datasets from the database of our company.
In order to protect personal privacy, we randomly replace the last two characters with another two characters for each sample \footnote{Sufficient data protection was carried out during the process of experiments to prevent the data leakage and the data was destroyed after the experiments were finished.
The data is only used for academic research and sampled from the original data, therefore it does not represent any real business situation in XXXX XXXX Services Group.}.

Details related to the dataset are shown in Table \ref{tab:dataset}.
Mac address is used to identify the physical address of hardware, and is composed of 16 hexadecimal numbers in which ``:" will be inserted after every two hexadecimal numbers.
The International Mobile Equipment Identity (IMEI) is used to identify mobile phones and satellite phones, and is made up of 15 or 17 numbers.
The ip address is made up of four parts which are separated by ``.", and each part ranges from 1 to 255.
The invoice code and invoice number are given by the tax department. The length of invoice code is 12 or 10 and the length of invoice number is 8. Both of them are composed of numbers.
The Chinese Mobile number consists of 11 numbers.
House id whose prefix is always ``17" and ``18"  is utilized to identify the house property, and is composed of 18 numbers.
The Car Engine number is exploited to identify a car's engine and contains different lengths of characters.
In China, company unicode is given by the administration for industry and commerce when a company is registered, and is made up of 18 characters.
Chinese Certificate number contains 18 characters and is used to identify the legal residents of China. The first 17 characters are numbers and the last character can either be number or ``X".
Car license is utilized to identify a car and is prefixed by a Chinese character which denotes a province of China. The rest of the characters in a car license are either numbers or upper case letters.
There are several suffixes of email address including ``qq.com", "163.com", "gmail.com", etc.
The lengths of bankcard numbers are different. However, for the same category of bankcard, the prefix is invariant.

For each dataset, we randomly select 10000 samples and combine them to construct the test set, which has 130000 items in total.
The rest of data is utilized as a train set.
Limited by the efficiency of the GP algorithm, we only sample a part of items from the train set for the training process. With regard to our method, except for the data chosen for training GP algorithm, the rest of data in the train set is used for training BPE.

In experiments, the same hyperparameters of our method and the baseline are equally defined.
For each kind of data, we randomly sample 2000 positive samples from it, and take the same number of negative samples from the remaining categories.
The training epoch size and the initialized population size are set as 5000 and 1000, respectively.
The minimum population size of our method is defined as 200.
To accelerate the training process of our method, the exponential decay parameter is set as 0.97.
The granularity of frequent items in BPE is controlled based on a threshold, which is equal to 0.02.
We begin to do divide-and-conquer after 30 training epochs and the precision threshold is set as 0.9.

\subsection{Strong Baseline}
To demonstrate the effectiveness of our method, a strong baseline in \cite{Bartoli2016Inference} is chosen for comparison, which achieves the state-of-the-art performance in the task of generating regular expression from  samples. The baseline is also built based on a genetic algorithm and generates regular expression character by character. In each training epoch of the baseline,  the population size is constant which will be demonstrated to be inefficient in the following subsection.

\subsection{Evaluation Metrics}
The results of experiments are reported based on precision, recall and F1 score. Given $n$ classes $C = {c_1, c_2, \cdots, c_n}$, for the $i_{th}$ category, the metrics are defined as follows.
\begin{equation}
    \begin{split}
        P_i = \frac{TP_i}{TP_i + FP_i} \\
        R_i = \frac{TP_i}{TP_i + FN_i}\\
        F1_i = \frac{2*P_i*R_i}{P_i+R_i}
    \end{split}
    \label{metrics}
\end{equation}
where $P_i$ , $R_i$ and $F1_i$ stand for precision, recall and F1 score, respectively. Besides, with related to the $i_{th}$ category, $TP_i$, $FP_i$ and $FN_i$ denote the number of true positives, false positives and false negatives , respectively.

\begin{table*}[]
\small
\renewcommand{\arraystretch}{1.85}
    \centering
    \begin{tabular}{cccccp{8cm}}
         \hline
         \textbf{Data Type} & \textbf{Algorithm}  & \textbf{Precision}  & \textbf{Recall} &  \textbf{F-score} & \textbf{Regular Expressions}\\
         \hline
         \multirow{2}{*}{Mac Address} &
           Baseline & 1.0000 & 0.9992 & 0.9995  &   \textbackslash w\textbackslash w:[\^{}\_]++  \\

           & Our method & \textbf{1.0000} & \textbf{1.0000} & \textbf{1.0000} &   [\^{}\textvisiblespace]8:[\^{}>]++|\textbackslash d:[\^{}\#]++|[\^{}:]++[\^{}\textvisiblespace]++                                            \\
         \hline
         \multirow{2}{*}{IMEI} &
           Baseline & 0.4889 & \textbf{0.9671} & 0.6495  &  (?<!\textbackslash w)\textbackslash d\textbackslash d(?:\textbackslash w\textbackslash w\textbackslash d\textbackslash d\textbackslash w\textbackslash d\textbackslash d)++\textbackslash w\textbackslash w\textbackslash d\textbackslash w\textbackslash w \textbackslash d                                            \\

           & Our method & \textbf{0.9648} & 0.9418 & \textbf{0.9531}  & \makecell[l]{
          8686\textbackslash d*+|86\textbackslash d\textbackslash d\textbackslash d\textbackslash d\textbackslash d\textbackslash d\textbackslash d8686\textbackslash d\textbackslash d|86\textbackslash d\textbackslash d86\textbackslash d\textbackslash d86\textbackslash d\textbackslash d++|86\textbackslash d\textbackslash d\textbackslash d\textbackslash d\textbackslash d\textbackslash d\textbackslash \\ d\textbackslash d8686 \textbackslash d|86\textbackslash d\textbackslash d\textbackslash d\textbackslash d\textbackslash d\textbackslash d\textbackslash d\textbackslash d\textbackslash d8686|86\textbackslash d\textbackslash d\textbackslash d\textbackslash d\textbackslash d\textbackslash d\textbackslash d\textbackslash d\textbackslash d86\textbackslash d\textbackslash  d \\ |86\textbackslash d\textbackslash d\textbackslash d\textbackslash d\textbackslash d\textbackslash d\textbackslash d86\textbackslash d\textbackslash d\textbackslash d\textbackslash d |\textbackslash d++[A-Za-z]\textbackslash d[A-Za-z]\textbackslash w\textbackslash d\textbackslash w++|......}
           \\
         \hline
         \multirow{2}{*}{IP Address} &
           Baseline & \textbf{1.0000} & 1.0000 & \textbf{1.0000} & \textbackslash d++\textbackslash.\textbackslash d[\^{}\_]++       \\

           & Our method & 0.9998 & 1.0000 & 0.9999 & \textbackslash .[\^{}\textvisiblespace]++|\textbackslash d++\textbackslash .[\^{}\textvisiblespace]++ \\
         \hline
         \multirow{2}{*}{Invoice Code} &
           Baseline & 0.8494 & 0.9675 & 0.9046 &    (?<!\textbackslash w)\textbackslash w\textbackslash w\textbackslash d[0-8]\textbackslash w(\textbackslash d\textbackslash w\textbackslash w\textbackslash w)++[0-8]\textbackslash w\textbackslash w
            \\

           & Our method & \textbf{0.9617} & \textbf{0.9956} & \textbf{0.9784} &
\makecell[l]{110019\textbackslash d*+|\textbackslash d10019\textbackslash d++|\textbackslash d110019\textbackslash d++|\textbackslash d\textbackslash d0019\textbackslash d\textbackslash d\textbackslash d\textbackslash d|\textbackslash d\textbackslash d001\textbackslash d\textbackslash d\textbackslash d\textbackslash d\textbackslash d \\ |\textbackslash \d\textbackslash d\textbackslash d 001\textbackslash d001\textbackslash d\textbackslash d|\textbackslash d\textbackslash d03\textbackslash d\textbackslash d\textbackslash d\textbackslash d\textbackslash d\textbackslash d|\textbackslash d\textbackslash d10019\textbackslash d\textbackslash w*+|\textbackslash d\textbackslash d\textbackslash d001\textbackslash d\textbackslash d\textbackslash d\textbackslash \\ d\textbackslash d\textbackslash d|......} \\
         \hline
         \multirow{2}{*}{Invoice Number} &
           Baseline & \textbf{0.9353} & 1.0000 & \textbf{0.9666} &    [0-8]\textbackslash d\textbackslash d\textbackslash d\textbackslash d\textbackslash d\textbackslash d\textbackslash d          \\

           & Our method & 0.9327 & 1.0000 & 0.9652 &
\textbackslash d\textbackslash d\textbackslash d\textbackslash d\textbackslash d\textbackslash d\textbackslash d\textbackslash d
            \\
         \hline
         \multirow{2}{*}{Mobile Number} &
           Baseline & 0.7452 & 1.0000 & 0.8540 &   (?<!\textbackslash d)\textbackslash d\textbackslash d\textbackslash d\textbackslash d\textbackslash d\textbackslash d\textbackslash d\textbackslash d\textbackslash d\textbackslash d\textbackslash d(?!\textbackslash w) \\

           & Our method & \textbf{0.7479} & 1.0000 & \textbf{0.8558}  &
\makecell[l]{13\textbackslash d1313\textbackslash d\textbackslash d++|13\textbackslash d15\textbackslash d*+|13\textbackslash d13\textbackslash d\textbackslash d13\textbackslash d\textbackslash d|13\textbackslash d13\textbackslash d\textbackslash d\textbackslash d13\textbackslash d|13\textbackslash d13\textbackslash d\textbackslash d\textbackslash \\ d\textbackslash d13|13 \textbackslash d13\textbackslash d13\textbackslash d\textbackslash d\textbackslash d|1315\textbackslash d++|13\textbackslash d\textbackslash d15\textbackslash d++|13\textbackslash d13\textbackslash \d\textbackslash d\textbackslash d\textbackslash d\textbackslash \d\textbackslash d|13\textbackslash d\textbackslash d\textbackslash \\ d13\textbackslash d\textbackslash d13|13\textbackslash d\textbackslash d \textbackslash d\textbackslash d13\textbackslash d*+|1313\textbackslash d*+|13\textbackslash d\textbackslash d\textbackslash d\textbackslash d\textbackslash d13\textbackslash d\textbackslash d|1513\textbackslash \\ d\textbackslash d++|13\textbackslash d\textbackslash d\textbackslash d13\textbackslash d13\textbackslash d|...... }\\
         \hline
         \multirow{2}{*}{House id} &
           Baseline & 0.9390 & 0.9005 & 0.9194  &    \makecell[l]{ [0-8]\textbackslash d[0-8]\textbackslash d[0-8]\textbackslash d[0-8][0-8][0-8]\textbackslash d\textbackslash d\textbackslash d\textbackslash d\textbackslash d\textbackslash d\textbackslash d\textbackslash d\textbackslash d|\textbackslash d\textbackslash d\textbackslash d\textbackslash d\textbackslash d\textbackslash d\textbackslash d\textbackslash d\textbackslash d\textbackslash \\ d[\^{}0-8]\textbackslash d\textbackslash d\textbackslash d[\^{} 0-8]\textbackslash d[\^{}0-8][\^{}0-8]|\textbackslash d\textbackslash d\textbackslash d\textbackslash d\textbackslash d\textbackslash d\textbackslash d\textbackslash d\textbackslash d\textbackslash d[\^{}0-8]\textbackslash d\textbackslash \\ d\textbackslash d[\^{}0-8]\textbackslash d[\^{}0-8]\textbackslash d                                        }
           \\
           & Our method & \textbf{0.9970} & \textbf{1.0000} & \textbf{0.9985} &
\makecell[l]{170\textbackslash d17\textbackslash d*+|170\textbackslash d\textbackslash d\textbackslash d170\textbackslash d++|170\textbackslash d\textbackslash d\textbackslash d\textbackslash d\textbackslash d\textbackslash d\textbackslash d(?:\textbackslash d\textbackslash d)++|\\ 17\textbackslash d\textbackslash d17\textbackslash w++|17\textbackslash d\textbackslash d\textbackslash d \textbackslash d17\textbackslash d\textbackslash d++|17\textbackslash d\textbackslash d\textbackslash d\textbackslash d\textbackslash d(?:\textbackslash d\textbackslash \\ d\textbackslash d\textbackslash d)++\textbackslash d++|\textbackslash d80\textbackslash d\textbackslash d80\textbackslash d++|\textbackslash d80\textbackslash d\textbackslash d\textbackslash d\textbackslash d\textbackslash d\textbackslash d8080\textbackslash d \textbackslash d\textbackslash d\textbackslash d\textbackslash d|......}
           \\
         \hline
         \multirow{2}{*}{Car Engine Number} &
           Baseline & 0.4643 & \textbf{0.9840} & 0.6309  & \makecell[l]{
(?:\textbackslash d\textbackslash d\textbackslash d\textbackslash d\textbackslash d\textbackslash d\textbackslash d\textbackslash d\textbackslash d\textbackslash d\textbackslash w)?+(?<![\^{},])\textbackslash w\textbackslash w\textbackslash w\textbackslash w(?:\textbackslash d*+\textbackslash w?+ \\\textbackslash w?+\textbackslash d*+[\^{}:][\^{}:][\^{},] [\^{}<][\^{}:]++)?+[\^{},]++}
           \\

           & Our method & \textbf{0.9416} & 0.9375 & \textbf{0.9396} & \makecell[l]{
[A-Za-z]\textbackslash d*+[\^{}']\textbackslash d*+|[\^{}\#]\textbackslash d[\^{}\#][\^{}\#]\textbackslash d\textbackslash d[\^{}\#]|\textbackslash d(\textbackslash w)|\textbackslash d(?:[\^{}\textbackslash d][\^{} \\ \textbackslash d])*+[\^{}\textbackslash d]\textbackslash d*+|[\^{}\textvisiblespace] [\^{}\textvisiblespace][\^{}\textvisiblespace][\^{}\textvisiblespace]\textbackslash d\textbackslash d\textbackslash d\textbackslash d\textbackslash d|......}
            \\
         \hline
         \multirow{2}{*}{Company Unicode} &
           Baseline & 0.9036 & 0.8888 & 0.8961  &  (?<!\textbackslash w)\textbackslash d\textbackslash d\textbackslash d\textbackslash d\textbackslash d\textbackslash d\textbackslash d++\textbackslash w++ \\

           & Our method & \textbf{0.9805} & \textbf{0.9728} & \textbf{0.9767} & \makecell[l]{
91(?:[\^{}"]\textbackslash d)++|\textbackslash d++MA\textbackslash w*+|91(?:\textbackslash d\textbackslash d)++\textbackslash w\textbackslash w|91\textbackslash d*+\textbackslash w\textbackslash w++|\textbackslash d++ \\\textbackslash w\textbackslash d++\textbackslash w|(?:\textbackslash d \textbackslash d)++[\textbackslash w]\textbackslash d|\textbackslash d++\textbackslash w\textbackslash w\textbackslash d++\textbackslash w|......}
            \\
         \hline
         \multirow{2}{*}{Chinese Certificate Number} &
           Baseline & 0.4480 & 0.9109 & 0.6006 & \textbackslash d\textbackslash d\textbackslash d\textbackslash d\textbackslash d\textbackslash d\textbackslash d\textbackslash d\textbackslash d\textbackslash d\textbackslash d\textbackslash d\textbackslash d\textbackslash d\textbackslash d\textbackslash d\textbackslash d\textbackslash w  \\

           & Our method & \textbf{0.8969} & \textbf{0.9831} & \textbf{0.9380} & \makecell[l]{
220\textbackslash d\textbackslash d\textbackslash d\textbackslash d\textbackslash d\textbackslash w++|\textbackslash d1010\textbackslash d\textbackslash d\textbackslash d\textbackslash d\textbackslash d\textbackslash w++|\textbackslash d10\textbackslash d++[\textbackslash w]|\textbackslash \\\d3030\textbackslash w*+|\textbackslash d30\textbackslash d++[\^{} \textvisiblespace] |\textbackslash d20\textbackslash d++[\textbackslash w]|\textbackslash d30\textbackslash d30\textbackslash d++|\textbackslash \d2020\textbackslash d++|\textbackslash d2030\textbackslash d++ \\
|\textbackslash d10\textbackslash d\textbackslash d\textbackslash d\textbackslash  \d\textbackslash d\textbackslash d\textbackslash d1030\textbackslash d\textbackslash d\textbackslash d \textbackslash d|\textbackslash d10\textbackslash d30\textbackslash d++|\textbackslash d3010\textbackslash d++|\textbackslash d110\textbackslash d++[\^{}/]|\textbackslash \\ d30\textbackslash d\textbackslash d\textbackslash d\textbackslash d\textbackslash d\textbackslash d\textbackslash d\textbackslash d30\textbackslash d\textbackslash d\textbackslash d\textbackslash d\textbackslash d|......}
           \\
         \hline
         \multirow{2}{*}{Car License} &
           Baseline & 0.9998 & 0.9998 & 0.9998 & (?<![\^{}\_])[\^{}\textbackslash w]\textbackslash w++  \\

           & Our method & \textbf{0.9999} & \textbf{1.0000} & \textbf{0.9999} &
[\^{}\textbackslash w][\^{}\textvisiblespace]++
           \\
         \hline
         \multirow{2}{*}{Email} &
           Baseline & 1.0000 & 1.0000 & 1.0000 & [\^{}@]++@[\^{}\_]++  \\

           & Our method & 1.0000 & 1.0000 & 1.0000 &
[\^{}@]*+@163\textbackslash .com|[\^{}@]*+@[\^{}"]*+
            \\
         \hline
         \multirow{2}{*}{Bankcard Number} &
           Baseline & 0.8897 & 0.9150 & 0.9022  & \textbackslash w\textbackslash w\textbackslash w\textbackslash d\textbackslash d\textbackslash d\textbackslash d\textbackslash w\textbackslash d\textbackslash w\textbackslash d\textbackslash d\textbackslash w\textbackslash d\textbackslash w\textbackslash d\textbackslash d\textbackslash d\textbackslash d++ \\

           & Our method & \textbf{0.9667} & \textbf{0.9942} & \textbf{0.9802} & \makecell[l]{
           622(?:\textbackslash d\textbackslash d)++|621(?:\textbackslash d\textbackslash d)++|622\textbackslash d\textbackslash d\textbackslash d\textbackslash d\textbackslash d\textbackslash d\textbackslash d\textbackslash d\textbackslash d\textbackslash d\textbackslash d\textbackslash d\textbackslash d|621\textbackslash \\
               d\textbackslash d\textbackslash d\textbackslash d\textbackslash d\textbackslash d\textbackslash d621\textbackslash d \textbackslash d++|621\textbackslash d\textbackslash d\textbackslash d\textbackslash d\textbackslash d\textbackslash d\textbackslash d\textbackslash d\textbackslash d\textbackslash d\textbackslash d\textbackslash d\textbackslash d|62\textbackslash d\textbackslash d\textbackslash
               \\
               d\textbackslash d|\textbackslash d++\textbackslash *++\textbackslash d++|......}
           \\
         \hline
    \end{tabular}
    \caption{The result of our method  and the baseline. Due to utilizing divide and conquer and BPE, some generated regular expressions of our methods are too long to be written here. Hence, we utilize "......" to indicate other possible regular expressions.}
    \label{tab:result}
\end{table*}

\begin{table*}[ht]
\linespread{1.5}
\renewcommand{\arraystretch}{1.5}
\centering
\begin{tabular}{lcccccc}
 \hline
  Positive & Negative  & Algorithm   & Init Pop Size  & Min Pop Size & Damping Index &  Elapsed time \\
 \hline
  \multirow{2}{*}{400}  & \multirow{2}{*}{400}  & Baseline & 400  & - & - &  0h25m40s \\

 &  & Our method & 400  & 100 & 0.99 &  \textbf{0h0m13s} \\
 \hline

 \multirow{2}{*}{1000}  & \multirow{2}{*}{1000}
   & Baseline & 1000  & - & - &  1h24m13s \\

  &  &   Our method & 1000  & 200 & 0.99 &  \textbf{0h0m58s} \\
 \hline

 \multirow{2}{*}{4000}  & \multirow{2}{*}{4000}
    & Baseline & 2000  & - & - &  > 1day \\

   &  &    Our method & 2000  & 200 & 0.99 &  \textbf{0h9m12s} \\
  \hline

  \multirow{2}{*}{10000}  & \multirow{2}{*}{10000}  & Baseline & 1000  & - & - &  >5days \\

  &  &    Our method & 2000  & 200 & 0.995 &  \textbf{0h15m13s} \\
 \hline
\end{tabular}
\caption{The training time of our method and the baseline.}
\label{tab:time_consuming}
\end{table*}

\subsection{Experimental Result}
Our method together with the baseline is evaluated on the test dataset mentioned above, which contains 13 categories of data and 130000 samples in total. The results are reported in Table \ref{tab:result} and demonstrate the effectiveness of our method, which outperforms the baseline on 10 kinds of dataset and achieves nearly 9.7 percent improvement on F1 score on average.
Especially, for these data with obvious frequent items, our method surpasses the baseline to a great extent.
For instance, our method outperforms the baseline by nearly 33 percent on the Chinese certificate number.
We have introduced the structure of the Chinese certificate number in the previous section.
In a specific Chinese certificate number, the first 2 characters stands for the province code, e.g. ``22" is the code of JiLin province of China. However, the baseline utilizes the pattern of  ``\textbackslash d\textbackslash d"  to capture these frequent items, which can be easily cheated by some adversarial samples. In the test set, the generated regular expression of the baseline will also match the bankcard number and the house id, which leads to the poor performance of the baseline.
Our method applies BPE to capture these frequent items and construct the pattern much more precisely, which makes our method more robust to deal with the adversarial samples.
For samples without frequent items, the performance of our method is similar to the baseline's performance, which means our modification is reasonable and valid.

In this work, we propose some tricks to accelerate the training process of the GP algorithm in order to make auto regular expression generation feasible for   industry applications.
To evaluate the efficiency of our method, we choose different numbers of positive and negative samples, and test our method and the baseline in the same running environment. The results are recorded in \ref{tab:time_consuming}.
Apparently, our method is much faster than the baseline, which is nearly 100 times faster than the baseline. When choosing thousands of training samples,  the baseline can hardly be trained.

In general, the results mentioned above demonstrate both effectiveness and efficiency of our method.

\begin{figure*}[t]
  	\begin{minipage}[t]{.4\textwidth}
  		\centering
      \includegraphics[width=\columnwidth]{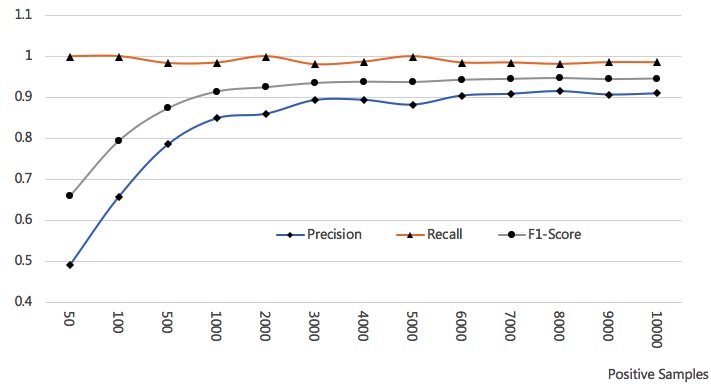}
  	\end{minipage}
  	\quad
  	\begin{minipage}[t]{.4\textwidth}
  		\centering
      \includegraphics[width=\columnwidth]{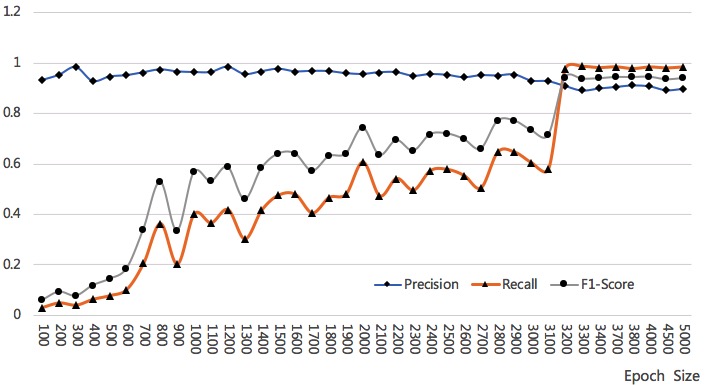}
  	\end{minipage}
  	\begin{minipage}[t]{.4\textwidth}
  		\centering
  		(a)
  	\end{minipage}
  	\quad
  	\begin{minipage}[t]{.4\textwidth}
  		\centering
  		(b)
  	\end{minipage}
  	\begin{minipage}[t]{.4\textwidth}
  		\centering
      \includegraphics[width=\columnwidth]{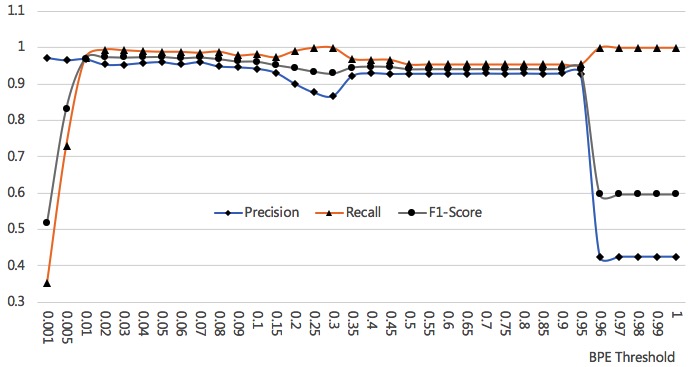}
  	\end{minipage}
  	\quad
  	\begin{minipage}[t]{.4\textwidth}
  		\centering
      \includegraphics[width=\columnwidth]{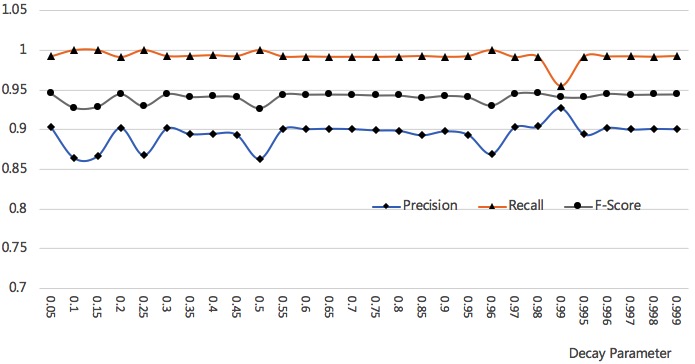}
  	\end{minipage}
  	\begin{minipage}[t]{.4\textwidth}
  		\centering
  		(c)
  	\end{minipage}
  	\quad
  	\begin{minipage}[t]{.4\textwidth}
  		\centering
  		(d)
  	\end{minipage}
    \caption{Hyperparamters analysis results. (a) , (b), (c), (d) represent the result of numbers of positive training samples, epoch size, BPE threshold and  decay parameter on population size, respectively.}
    \label{fig:chaocan}
\end{figure*}

\subsection{Hyperparameter Analysis}
In this subsection, we analyze some vital hyperparameters including the number of positive training samples, epoch size, exponential decay parameter on population size and the threshold of BPE, which may have effects on the performance of our method.
Besides, we only report the performance of our method on the Chinese certificate number since it is the most challenging among all dataset.
For a specific hyperparameter, the rest of experiment settings are kept unchanged.
Figure \ref{fig:chaocan} illustrates the results of these hyperparameters, which is reported according to the metrics defined in Equation \ref{metrics}. The square, triangle and dot represent precision, recall and F1 score, respectively.

\textbf{Numbers of Positive Training Samples}
In hyperparameter analysis, the number of positive training samples and the number of negative training samples are equal.
Surprisingly, there is no significant change in the recall.
We argue that more training samples will make our method generate more specific regular expressions.
At the beginning, the generated regular expression is too generalized to distinguish positive samples and negative samples due to the limited training samples, which leads to low precision and high recall.
As the training samples grow, the precision and F1 score of our method are improved.
In general, the performance is sensitive to the number of training samples at the beginning, and the upswing slows down when the number reaches to 3000. For efficiency, we suggest the number of the positive training samples can be set between 3000 and 10000.

\textbf{Epoch Size}
The corresponding curves of recall and F1 score rise with fluctuation, which is as expected since more and more outstanding genes can be saved and generated with the epoch size growing.
In addition, our algorithm utilizes divide-and-conquer strategy during each epoch. As the epoch size grows, more and more regular expressions will be generated for these mismatched samples in the last round of training epoch.
The precision has no significant change since we utilize BPE to improve the ability of regular expressions to distinguish the positive samples and negative samples.
It seems that bigger epoch size will help to generate better regular expressions. However, once the epoch size reaches 3300, the performance will not be improved. In general, for our method, the epoch size can be set as 3300 to achieve the best performance.

\textbf{BPE Threshold}
As shown in (c), the recall and F-score are pool when the threshold is set as 0.001.
In our BPE algorithm, the smaller threshold is the more frequent items will be captured.
However, if the threshold is too small, our BPE algorithm will suffer from overfitting since the whole training sample may be recognized as a frequent item.
Besides, the bigger threshold will generate the items with higher frequency.
However, if the threshold is big enough,  the BPE algorithm can hardly capture any frequent item and our method will rollback to the character-based method. Hence, the curve of F1 score grows first and then tends to be stable, and goes down in the end.

\textbf{Decay Parameter}
From (d) of Figure \ref{fig:chaocan}, we find that the exponential decay parameter on population size does not affect the performance of our method. Some random strategies during the initialization stage of our method can account for the fluctuations in the curve. The results suggest that doing  exponential decay parameter on population size is reasonable and valid.

\section{conclusion}
Traditional regular expression generation tasks are divided into two categories including generating regular expression from natural language and generating regular expression from samples.
In this work, we focus on the second kind of task. 
For this task, although some effective methods are proposed, these methods suffer from inefficiency and can hardly be exploited to model industry applications. 
In order to make automatic regular expression generation available for industry, we propose a novel genetic algorithm which is motivated by \cite{Bartoli2016Inference}.
During the initialized stage, we generate some candidate regular expressions by using syntactic trees. Based on a pre-defined fitness function, the genetic mutation operation and crossover operation are carried out to find result with the best fitness from candidates.
Different from the character-based methods, our method first utilizes BPE to extract some frequent items from the training examples, which are then utilized to construct more specific regular expressions. In order to accelerate training procedure, we conduct exponential decay on the number of candidates during each epoch.

Our method and a strong baseline are tested on 13 categories of data. The result indicates the validity of our method, which outperforms the baseline by nearly 9.7 percent on average.
Especially, for those data with obvious frequent item like Chinese certificate number, our method even achieves 33 percent improvement. Furthermore, by doing exponential decay, our method is nearly 100 times faster than the baseline.

The future work will concentrate on the improvement of mutation and crossover operations so that more effective genes can be generated. In addition, we find that the generated regular expressions of our method are of great complexity if BPE is utilized. Hence, we will explore some methods to simplify the output of our algorithm.

\bibliographystyle{ACM-Reference-Format}
\bibliography{Revisiting_Regex_Generation_for_Modeling_Industrial_Applications_by_Incorporating_Byte_Pair_Encoder}

\end{document}